\documentclass{article}

\usepackage[main, preprint]{neurips_2026}

\usepackage[utf8]{inputenc} 
\usepackage[T1]{fontenc}    
\usepackage{hyperref}       
\usepackage{url}            
\usepackage{booktabs}       
\usepackage{amsfonts}       
\usepackage{nicefrac}       
\usepackage{microtype}      
\usepackage{xcolor}         
\usepackage{colortbl} 
\usepackage{wrapfig}

\usepackage{amsmath}
\usepackage{amssymb}
\usepackage{amsfonts}
\usepackage{amsthm}

\theoremstyle{definition}

\theoremstyle{remark}
\title{PH-Dreamer: A Physics-Driven World Model via Port-Hamiltonian Generative Dynamics}

\usepackage{multirow}
\usepackage{microtype}
\usepackage{graphicx}
\usepackage{subcaption}
\usepackage{booktabs} 
\usepackage{hyperref}

\usepackage{hyperref}
\usepackage{pifont}
\usepackage{amsmath}
\usepackage{amssymb}
\usepackage{mathtools}
\usepackage{amsthm}
\usepackage{svg}
\usepackage{float}
\usepackage[capitalize,noabbrev]{cleveref}

\author{%
  Xueyu Luan \\
  School of Electronic and Information Engineering \\
  Tongji University \\
  Shanghai 200092, China \\
  \texttt{2531946@tongji.edu.cn} \\
  \And
  Chenwei Shi \\
  Shanghai Research Institute for Intelligent Autonomous Systems \\
  Tongji University \\
  Shanghai 200092, China \\
  \texttt{2511973@tongji.edu.cn} \\
}

\begin{document}

\maketitle

\begin{abstract}
World models built on recurrent state space architectures enable efficient latent imagination, yet remain physically unstructured, producing dynamics that violate conservation and dissipative principles. We introduce a unified Port-Hamiltonian framework that remedies this through three synergistic mechanisms. First, we embed implicit physical priors into recurrent transitions by modeling projected latent evolution as action controlled energy routing governed by flow and dissipation, biasing the projected PH phase space toward a more compact and physically structured representation. Second, we develop a kinematics aware energy world model that estimates the Hamiltonian and power balance from proprioceptive observations, providing an explicit physical signal for thermodynamic reasoning. Third, leveraging these energy gradients, we establish an energy guided Actor-Critic that uses Lagrangian multipliers to regularize policy optimization toward lower energy and smoother control. Across visual control benchmarks, this paradigm not only attains superior asymptotic returns but also elevates internal simulator fidelity by establishing a tighter, lower variance alignment between imagined and real rewards, all while reducing latent phase space volume by 4.18–8.41\%, energy consumption by up to 7.80\%, and mean squared jerk by up to 9.38\%.

\end{abstract}

\section{Introduction}

Intelligent world modeling demands the internalization of fundamental physical regularities to ensure that latent transitions remain strictly grounded in the causal structure of the environment \citep{shang2025roboscape,li2025pin}. However, purely data driven generators like Sora \citep{brooks2024sora} often violate fundamental mechanics of gravity and fluid dynamics \citep{guo2025t2vphysbench}, illustrating that photorealism does not guarantee physical understanding. While hybrid frameworks attempt to inject structure via simulation \citep{zhou2024genesis, liu2024physgen} or PDE constraints \citep{bastek2025physics}, scaling alone cannot yield robust generalization in physics centric evaluations \citep{kang2025far, motamed2025generative, zheng2025vbench}. In model based reinforcement learning, Recurrent State Space Models (RSSMs) \citep{hafner2019learning, hafner2020dream, hafner2021mastering, hafner2025mastering} provide a natural substrate for action conditioned dynamics, yet remain unstructured. Learned end-to-end, these models offer no guarantees regarding energy consistency or dissipation mechanisms \citep{toth2020hamiltonian, de2018end, Lutter2019RP19}.

Physics-Informed Model-Based Reinforcement Learning attempts to bridge this gap by explicitly embedding known physical laws or analytical simulators into world models \citep{ramesh2023physics,yang2020learning}. While effective in improving physical fidelity for narrowly specified systems, such approaches typically depend on hand-crafted equations, domain-specific assumptions, or differentiable physics engines, leading to substantial modeling overhead and poor task generalization. As a result, explicit physics integration often sacrifices scalability and adaptability for fidelity, leaving unresolved the challenge of learning world models that are both physically grounded and broadly transferable.

Taken together, these limitations suggest that the prevailing reliance on static priors \citep{shi2023learning} or analytical constraints \citep{ramesh2023physics} remains insufficient. By reducing physical laws to closed-system conservation, current paradigms overlook the energy injection intrinsic to actuation. We contend that robust generalization demands a conceptual perspective: reframing physics not as a boundary condition to be preserved, but as the generative mechanism by which control actively remodels the system's energy landscape. This operationalizes our guiding insight: \textbf{Physics in world models is not a static conservation constraint, but a dynamic accounting of energy flow driven by control.}

Building on this energy centric insight, we propose a framework that unifies implicit and explicit Port Hamiltonian constraints. First, we introduce Port-Hamiltonian-inspired regularization on projected RSSM latents, framing latent transitions as energy-flow dynamics and empirically biasing the projected PH phase space toward a more compact and physically structured subspace. To anchor this compact geometry to physical reality, we develop an explicit energy world model that infers system energy from kinematic observations to provide a precise grounding for thermodynamic reasoning. This definitive anchor empowers the agent to transcend passive prediction. Empowered by this explicit energy world model, we further establish an energy-guided Lagrangian-regularized Actor-Critic framework. By using Lagrange multipliers to penalize learned estimates of action-induced energy variation and action-smoothness violations, we bias the policy toward smoother and more energy-efficient control in the evaluated domains.

\begin{itemize}
  \item We introduce a Port-Hamiltonian RSSM that incorporates implicit physical priors by regularizing projected latent dynamics in a structured phase space, thereby providing the architecture with a PH-inspired physical bias.
  \item We develop a Kinematics Aware Energy World Model to explicitly recover the Hamiltonian from proprioceptive observations and predict post action energy variations, providing a definitive physical anchor for reasoning.
  \item We establish an Energy-Guided Actor-Critic that uses Lagrange multipliers to encourage smoother and more energy-efficient control through learned physical regularizers.
\end{itemize}

\begin{figure*}[t]
  \centering
  \includegraphics[width=0.396\textwidth]{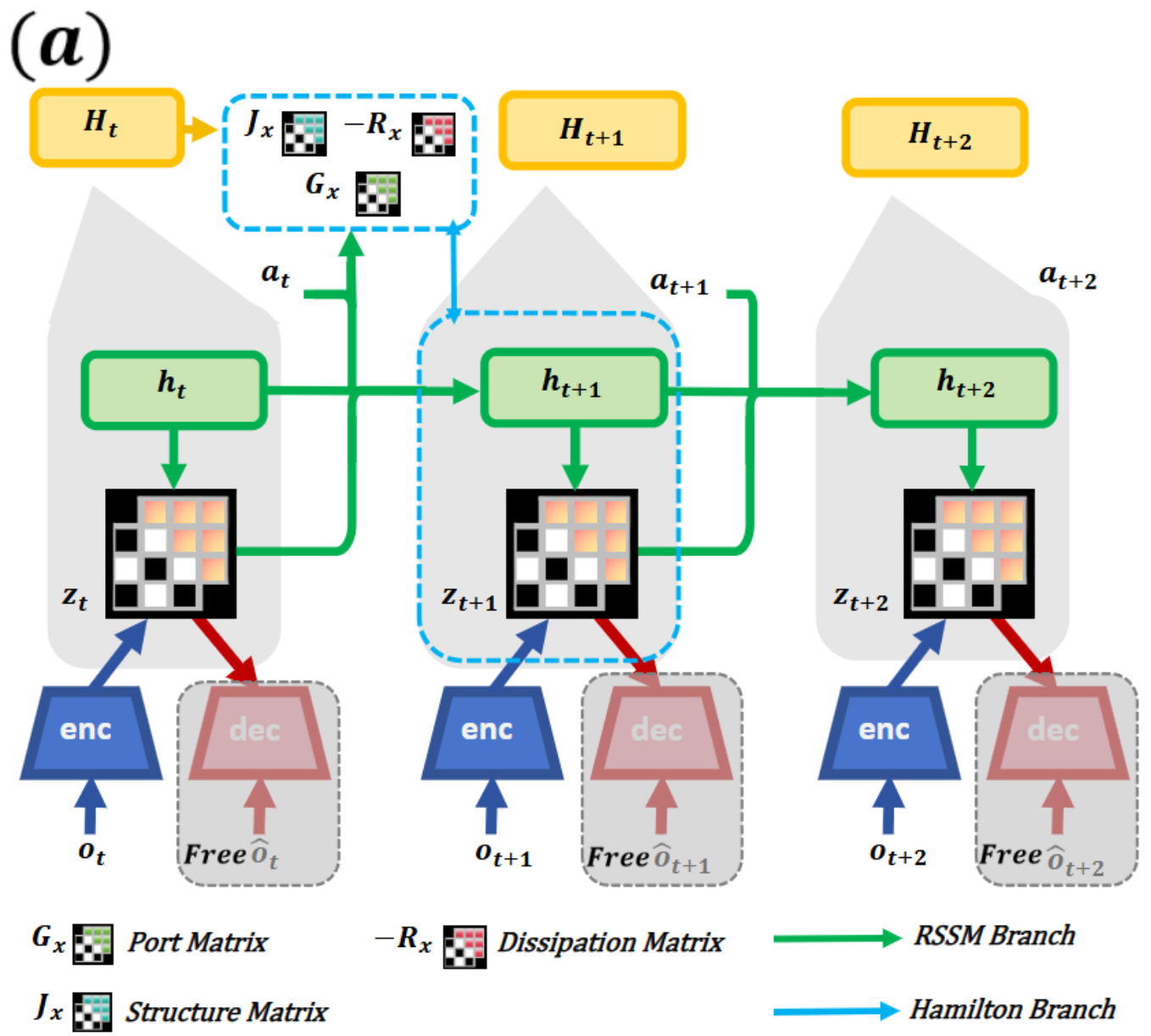}
  \includegraphics[width=0.484\textwidth]{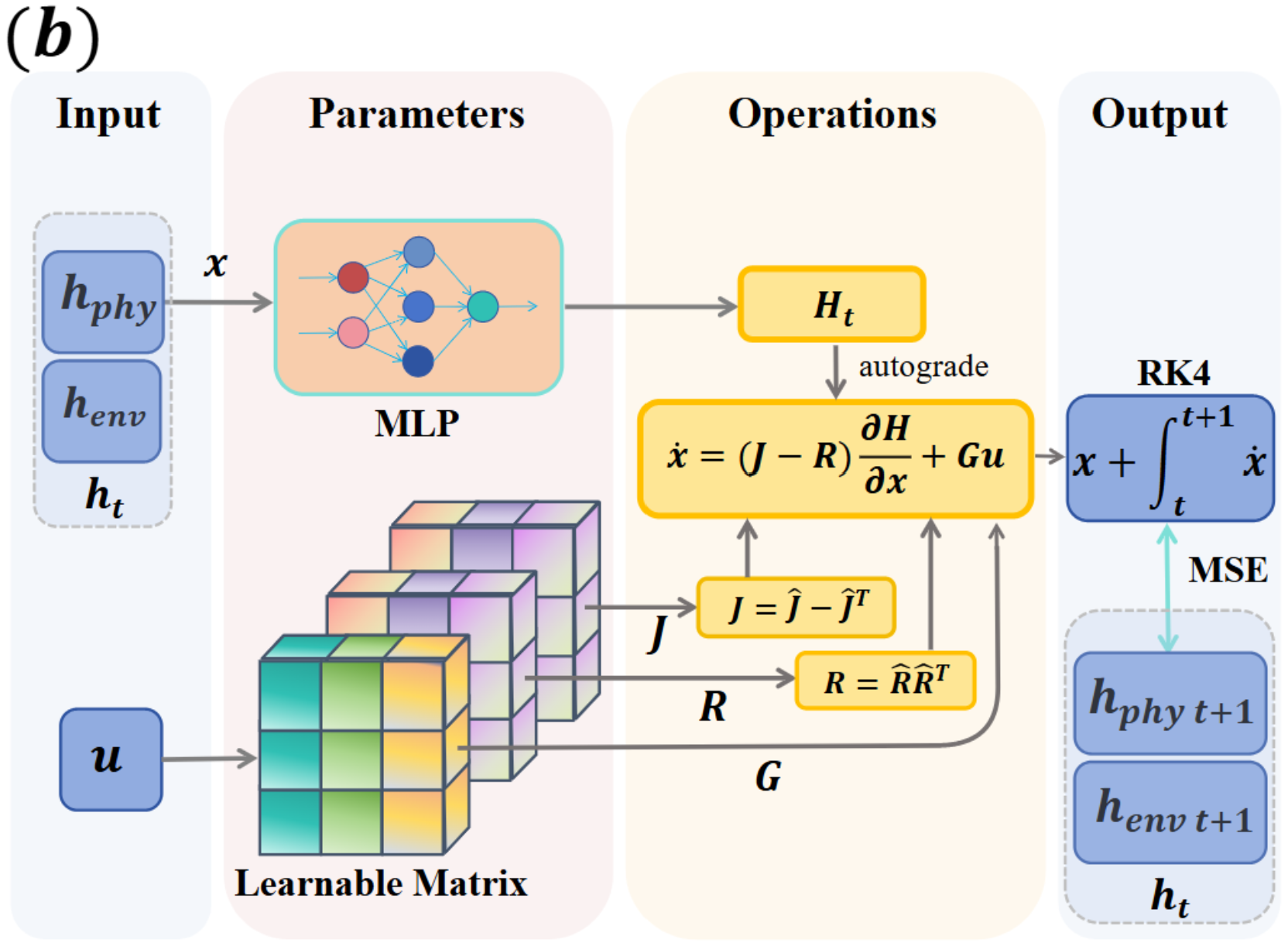}
  \includegraphics[width=0.88\textwidth]{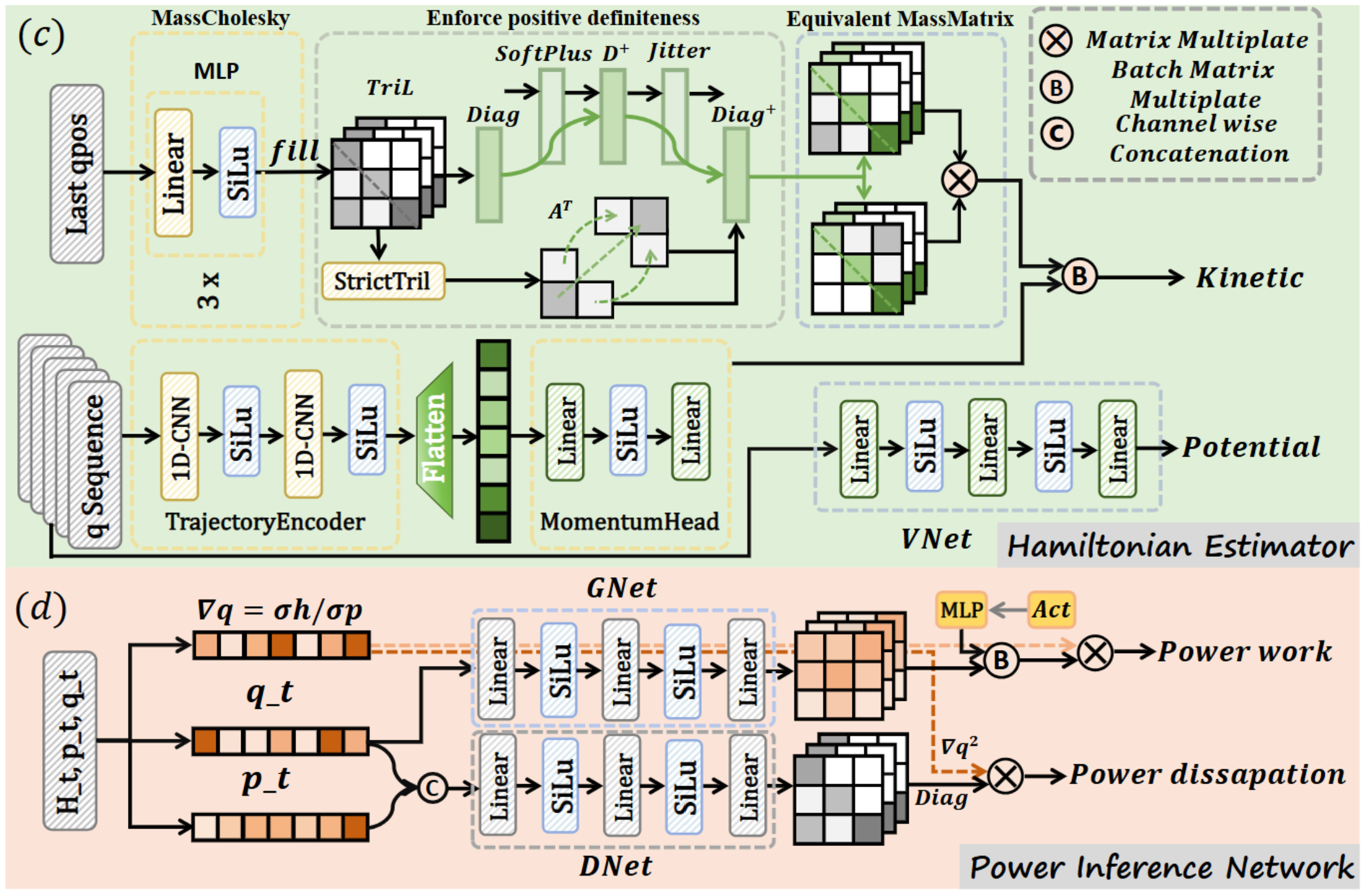}
  \caption{
    \textbf{Architecture of the Port Hamiltonian world model.} \textbf{(a)} Implicit structural constraints regularizing the recurrent state space transitions within the RSSM backbone. \textbf{(b)} Internal configuration of the implicit Hamiltonian estimator for unsupervised latent geometry supervision. \textbf{(c)} Network architecture for explicit Hamiltonian estimation grounded in proprioceptive kinematic observations. \textbf{(d)} Action driven energy inference capturing exogenous power injection and internal dissipation to facilitate energy guided policy optimization.
  }
  \label{OverallFrame}
  \vskip -0.1in
\end{figure*}

\section{Related Work}
\textbf{Latent World Models and Representation Learning.}
Latent dynamics models have established themselves as the cornerstone of sample-efficient reinforcement learning. The Recurrent State-Space Model (RSSM), established by PlaNet~\citep{hafner2019learning} and Dreamer~\citep{hafner2020dream}, provides a robust framework for learning compact representations from high-dimensional sensory inputs. Recently, DreamerV3~\citep{hafner2025mastering} demonstrated that RSSM-based agents could master diverse tasks with fixed hyperparameters, solidifying RSSM as the state-of-the-art backbone for online RL. While the latest DreamerV4~\citep{hafner2025training} explores Transformer-based architectures and flow matching to scale on large offline datasets, RSSM remains the preferred choice for online interaction due to its inference efficiency.

Driven by this success, numerous variants have emerged to enhance RSSM capabilities. Some replace the RNN backbone with Transformers (e.g., TransDreamer~\citep{dongare2025transdreamerv3}, STORM~\citep{zhang2023storm}) to capture long-term dependencies; others incorporate object-centric priors (e.g., SOLD~\citep{mosbach2025sold}, FOCUS~\citep{ferraro2025focus}) for scene decomposition to better capture key information, or modify objectives to enhance safety~\citep{huang2024safedreamer} and multi-task adaptation~\citep{ying2023task}. 
However, despite these functional improvements, these models remain fundamentally data-driven. They rely on statistical correlations to approximate transitions without embedding intrinsic physical laws. Consequently, they often struggle with \emph{physical consistency} in out-of-distribution scenarios—such as one-shot generalization or long-horizon energy conservation—where statistical patterns are insufficient~\citep{ai2025review}.
Departing from this purely statistical paradigm, our work introduces Port-Hamiltonian-inspired regularization into projected RSSM latents, encouraging the learned dynamics to respect conservation- and dissipation-like structure while retaining visual representational power.

\textbf{Physics-Informed Neural Dynamics.}
To enforce physical plausibility, foundational approaches like HNNs~\citep{greydanus2019hamiltonian} parameterize energy functions to ensure dynamics satisfy canonical equations. Such structure-preserving priors have been subsequently adapted for reinforcement learning (RL) to enhance sample efficiency~\citep{ramesh2023physics,levy2025learning}; however, these methods typically rely on low-dimensional states or explicit kinematic knowledge. Conversely, while Hamiltonian Generative Networks (HGNs)~\citep{toth2020hamiltonian} lift physical constraints into latent visual spaces, they remain restricted to passive, energy-conserving settings. Existing methods thus fail to simultaneously account for visual perception and the actuation-dissipation dynamics inherent in robotic control—a gap our Port-Hamiltonian framework specifically bridges.

\emph{Our Approach.} PH-RSSM synergizes implicit geometric regularization with explicit thermodynamic inference to connect high-dimensional visual perception with learned physical reasoning. We apply Port-Hamiltonian-inspired priors to projected latent transitions while simultaneously estimating energy representations from low-dimensional kinematic states. This dual route paradigm overcomes the restrictive assumptions of conservative closed systems \citep{toth2020hamiltonian} by accounting for action driven power flow and dissipation. By modeling the balance between work and energy, our framework provides a physically motivated regularization mechanism for reinforcement learning \citep{ramesh2023physics}, encouraging smoother and more energy-efficient control strategies. 

\section{Methods}
\label{sec:method}
We propose PH-RSSM to fundamentally recast latent evolution as a controlled physical process. Building upon the R2Dreamer architecture \citep{morihirar2}, Section~\ref{sec:PHRSSM} formalizes Port Hamiltonian world models by projecting dynamics into a structured phase space via curriculum geometric integration to enforce implicit physical consistency. Section~\ref{sec:energy_inference} introduces kinematics aware energy inference to explicitly recover system energy and power balance from proprioceptive observations. Finally, Section~\ref{sec:ac-learning} establishes an energy-guided actor-critic framework that leverages learned energy gradients to steer policies toward smoother and more energy-efficient control regimes. Figure~\ref{OverallFrame} illustrates the complete architecture.

\subsection{Port Hamiltonian World Models}
\label{sec:PHRSSM}

We formulate the environment as a POMDP where image observations $o_t$ and actions $a_t$ drive the latent state $s_t = (h_t, z_t)$, following:
\begin{equation}
    h_t = f_\phi(h_{t-1}, z_{t-1}, a_{t-1}), \quad z_t \sim q_\phi(z_t \mid h_t, \text{o}_t).
    \label{eq:rssm_recurrence}
\end{equation}
While R2Dreamer architectures successfully filter visual distractors, their reliance on physically agnostic GRUs fails to enforce energy conservation, often yielding implausible predictions during long temporal rollouts.

To rectify this, we partition the deterministic state into $h_t = [h_t^{\text{phys}}; h_t^{\text{env}}]$, where $h_t^{\text{phys}}$ distills features for primary physical modeling and $h_t^{\text{env}}$ captures residual environment dynamics. We project this physical subset into a low dimensional latent phase space $x_t \in \mathbb{R}^n$ via a learnable mapping:
\begin{equation}
    x_t = \phi_{\text{phys}}(h_t^{\text{phys}}) \triangleq \mathbf{W}_{\text{proj}} h_t^{\text{phys}} + b_{\text{proj}}.
    \label{eq:projection}
\end{equation}
This projection disentangles fundamental physical evolution from auxiliary environment semantics, facilitating stable Hamiltonian gradient $\nabla_x H$ computation while implicitly identifying the system generalized coordinates.

Rather than replacing the backbone, we utilize Port Hamiltonian (PH) dynamics as a structural regularizer, supervising latent geometry through a parallel shadow transition:
\begin{equation}
    \dot{x}_t = [\mathbf{J}(x_t) - \mathbf{R}(x_t)] \nabla_{x} H(x_t) + \mathbf{G}(x_t) a_t.
    \label{eq:ph_ode}
\end{equation}
Here, the algebraic parameterizations of $\mathbf{J}, \mathbf{R}, \mathbf{G}$ and $H$ impose a passivity-consistent energy-balance bias within the auxiliary projected PH dynamics. We compute physical predictions $x_{t+1}^{\text{pred}}$ via a Fourth Order Runge Kutta integration scheme for optimization only. Crucially, applying stop gradients to $h_t^{\rm phys}$ and $h_{t+1}^{\rm phys}$ compels the PH module to track existing semantic features without distorting the robust encoder outputs. Training optimizes a composite objective $\mathcal{L}_{\text{total}} = \mathcal{L}_{\text{RSSM}} + \lambda_{\text{PH}}(t) \cdot \mathcal{L}_{\text{PH}}$, where $\mathcal{L}_{\text{PH}} = \mathbb{E}_{t \sim \mathcal{D}} [ \| \text{sg}(\mathbf{x}_{t+1}^{\text{post}}) - \mathbf{x}_{t+1}^{\text{pred}} \|_2^2 ]$ updates only $\theta_{\rm PH}{=}\{\phi_{\rm phys},H_\phi,J_{\rm raw},R_{\rm raw},G\}$. To reconcile representation flexibility with structural rigidity, we propose Curriculum Geometric Integration, linearly annealing $\lambda_{\text{PH}}$ from 0 to $\lambda_{\max}$. This schedule allows the model to capture rich visual features before gradually introducing the physical regularizer, improving projected compactness and rollout stability in our experiments. 

\subsection{Kinematics Aware Energy Inference via Port-Hamiltonian Dynamics}
\label{sec:energy_inference}

To facilitate the energy-regularized policy optimization, we introduce a differentiable energy world model that infers the system's Hamiltonian (total energy) and its temporal evolution directly from proprioceptive kinematic states. Let $q_t \in \mathbb{R}^{d_q}$ denote the generalized coordinates (e.g., joint angles) at time $t$, and $a_t \in \mathbb{R}^{d_a}$ denote the applied action. 

Since the generalized momentum $p_t \in \mathbb{R}^{d_q}$ is often not explicitly provided by standard environmental observations, we first infer it from a brief kinematic history sequence $q_{t-k:t}$ (where $k$ is the history window size) using a temporal convolutional network: $p_t = f_{\theta_p}(q_{t-k:t})$. With the kinematic state $(q_t, p_t)$ established, we parameterize the system's Hamiltonian $\mathcal{H}_t$ as the sum of potential energy $V(q_t)$ and kinetic energy $K(q_t, p_t)$:
\begin{equation}
    \mathcal{H}(q_t, p_t) = V_{\theta_V}(q_t) + \frac{1}{2} p_t^\top M_{\theta_L}^{-1}(q_t) p_t
\end{equation}
where $V_{\theta_V}$ is a multi-layer perceptron (MLP) mapping $q_t \rightarrow \mathbb{R}$. To strictly guarantee the positive definiteness of the inverse mass matrix $M_{\theta_L}^{-1}(q_t)$, we utilize a Cholesky-like decomposition parameterized by a neural network. Specifically, we learn a lower triangular matrix $L_{\theta_L}(q_t) \in \mathbb{R}^{d_q \times d_q}$ with strictly positive diagonal elements enforced via a softplus activation, yielding $M_{\theta_L}^{-1}(q_t) = L_{\theta_L}(q_t) L_{\theta_L}(q_t)^\top$.

Crucially, to predict the energy variation induced by the agent's action without rolling out a full dynamics simulator, we exploit the power balance equation dictated by Port-Hamiltonian theory. The generalized velocity $\dot{q}$ is derived exactly through automatic differentiation of the estimated Hamiltonian: $\dot{q} = \nabla_{p_t} \mathcal{H}(q_t, p_t)$.

The continuous time derivative of the Hamiltonian is governed by the power injected into the system via external work ($\mathcal{P}_{\text{work}}$) minus the power dissipated by internal friction ($\mathcal{P}_{\text{diss}}$). We model these components using an action input port matrix $G_{\theta_G}(q_t) \in \mathbb{R}^{d_q \times d_a}$ and a state-dependent positive damping diagonal matrix $D_{\theta_D}(q_t, p_t) \in \mathbb{R}^{d_q \times d_q}_{\succ 0}$, let $\tilde{a}_t = \phi_{\theta_\phi}(a_t)$ denote the action after passing through a lightweight MLP. Then:

\begin{equation}
    \begin{aligned}
        \mathcal{P}_{\text{work}} &= \dot{q}^\top G_{\theta_G}(q_t) \tilde{a}_t \\
        \mathcal{P}_{\text{diss}} &= \dot{q}^\top D_{\theta_D}(q_t, p_t) \dot{q}
    \end{aligned}
\end{equation}
We discretize the induced $(q,p)$ PH dynamics with RK4:
\begin{equation}
    (q_{t+1},p_{t+1})=\operatorname{RK4}((f_q,f_p),(q_t,p_t),\tilde a_t,\Delta t),\quad
    \mathcal{H}_{t+1}=\mathcal{H}(q_{t+1},p_{t+1})
\end{equation}
By keeping the entire computational graph differentiable, this formulation provides an analytical energy gradient $\nabla_{a_t} \mathcal{H}_{t+1}$ with respect to the action, seamlessly paving the way for the downstream energy-regularized Actor-Critic learning.

\subsection{Actor-Critic Learning with Hamiltonian Constraints}
\label{sec:ac-learning}
Leveraging the explicit energy modeling from Section~\ref{sec:energy_inference}, we propose an Actor-Critic framework optimized via Lagrange multipliers. This formulation incorporates Hamiltonian energy and smoothness constraints, steering the policy to learn energy efficient and smooth action strategies without compromising task returns.

During the second stage of our training curriculum, we freeze the parameters of the world model components to exclusively optimize the actor and value networks. To ensure physical plausibility during this fine-tuning phase, we constrain the generated actions utilizing an external Hamiltonian energy world model. Given a generalized coordinate history of length $k$, an action vector $\mathbf{a}_t \in \mathbb{R}^{A}$, and a time step $\Delta t$, the model predicts the subsequent energy state as $\hat{H}_{t+1} = H_{\phi}(\mathbf{q}_{t-k+1:t}, \mathbf{a}_t, \Delta t)$. Over a batch of $N$ samples, we impose two critical physical constraints. First, we enforce an energy constraint $C_{\text{energy}}$ to penalize energy-violating action directions:
\begin{equation}
C_{\text{energy}} = \frac{1}{N}\sum_{i=1}^{N} \left( \nabla_{\mathbf a_i} H_{\phi}(q_i,a_i,\Delta t)^{\top}\mathbf a_i \right)^2 .
\end{equation}
We further introduce an \textbf{Action Smoothness Regularizer} $C_{\text{smooth}}$. This term measures the directional curvature of the learned Hamiltonian along the action direction through a Hessian-vector product. In the actor objective, it serves as a second-order energy-variation signal that discourages action directions associated with abrupt changes in the inferred energy landscape:
\begin{equation}
C_{\text{smooth}} = \frac{1}{N}\sum_{i=1}^{N} \mathbf a_i^{\top} \left(\nabla_{\mathbf a_i}^{2} H_{\phi}(q_i,a_i,\Delta t)\right) \mathbf a_i .
\end{equation}

Following the latent imagination paradigm, the primary component of our optimization objective uses the Dreamer-style imagination loss in our implementation, dedicated to maximizing task rewards based on imagined trajectories. Define continuation discount weights and normalized advantage as $w_t = \prod_{\tau=1}^{t}\left(\hat c_{\tau}\gamma\right)$ with $\gamma = 1-\frac{1}{\text{horizon}}$, and $\hat A_t = \frac{R_t^{\lambda}-V_t}{S_R}$ where $S_R=\max\!\left(1,\;Q^{\mathrm{EMA}}_{0.95}(R^\lambda)-Q^{\mathrm{EMA}}_{0.05}(R^\lambda)\right)$. The actor loss is evaluated as:
\begin{equation}
L_{\text{actor}}(\theta) = \mathbb E\!\left[ \sum_{t=1}^{H} \operatorname{sg}(w_t)\left( -\log \pi_{\theta}(a_t|s_t)\operatorname{sg}(\hat A_t) -\eta\,\mathcal H[\pi_{\theta}(\cdot|s_t)] \right) \right].
\end{equation}
Simultaneously, the critic is trained with both $\lambda$-return target and slow-value target:
\begin{equation}
L_{\text{critic}}(\psi) = \mathbb E\!\left[ \sum_{t=1}^{H} \operatorname{sg}(w_t)\left( -\log p_{\psi}(\tilde R_t^{\lambda}|s_t) -\log p_{\psi}(V_t^{\text{slow}}|s_t) \right) \right],
\end{equation}
where $\tilde R_t^{\lambda}$ denotes the padded $\lambda$-return target used in implementation. Here $s_t$ is the latent state, $a_t$ is the action, $R^{\lambda}$ is the $\lambda$-return, and sg denotes the stop-gradient operator.

These respective losses are aggregated using configuration coefficients $\alpha_p$ and $\alpha_v$ to form the composite actor-critic loss $L_{\text{AC}}=\alpha_p L_{\text{actor}}+\alpha_v L_{\text{critic}}$. The comprehensive optimization objective integrates this loss with the aforementioned physical constraints as $L_{\text{total}} = L_{\text{AC}} +\lambda_e(C_{\text{energy}}-\epsilon_e) +\lambda_s(C_{\text{smooth}}-\epsilon_s)$. Crucially, the primal loss omits an explicit hinge mechanism. Instead, constraint non-negativity is maintained by projected dual ascent: $\lambda_e \leftarrow \max\!\left(0,\lambda_e+\eta_{\lambda}(C_{\text{energy}}-\epsilon_e)\right)$ and $\lambda_s \leftarrow \max\!\left(0,\lambda_s+\eta_{\lambda}(C_{\text{smooth}}-\epsilon_s)\right)$. Through this dual projection strategy, the primal objective remains linear with respect to margin violations, yielding a physics aware finetuning procedure that preserves latent imagination updates while regulating energetic and kinematic properties of generated actions.

\section{Experiments}
\label{sec:experiments}

In this section, we conduct a series of experiments to validate the core claims of our work. Our evaluation is structured to answer the following key questions: 

\begin{itemize}
    \item \textbf{Latent Compactness and Physical Distillation:} Can the PH-inspired regularizers reduce non-physical variance, yielding a more compact projected PH phase space while maintaining or exceeding the task performance of unconstrained baselines?

    \item \textbf{Fidelity of Energy Prediction and Variation:} Can the explicit energy model estimate system energy and predict its temporal fluctuations from kinematic observations?
    
    \item \textbf{Energy Guided Policy Optimization: } Does the energy minimization constraint effectively steer the agent toward control strategies that are both energy efficient and kinematically smooth?
\end{itemize}

\subsection{Experimental Setup}
We evaluate our method on high dimensional visual control benchmarks designed to probe physical reasoning capabilities. For general continuous control, we utilize the DeepMind Control Suite~\citep{tassa2018deepmind}, selecting the \emph{Cheetah Run}, \emph{Reacher Easy}, \emph{Hopper Hop}, \emph{Walker Stand}, \emph{Walker Walk} and \emph{Walker Run} tasks. Notably, this suite provides accessible ground truth energy metrics, facilitating the rigorous verification of our learned Hamiltonian dynamics against physical reality. To systematically assess the efficacy of the energy informed policy learning, we formulated two customized evaluation metrics leveraging the low level APIs of MuJoCo, specifically quantifying the energetic efficiency and kinematic smoothness of the derived control policies.

\begin{table*}[t]
  \caption{\textbf{Bridging the Reality Gap.} Comparison of asymptotic evaluation performance and internal imagination fidelity. We report two coupled metrics: (1) the asymptotic performance in the true environment, measured as the best return obtained by deploying the fully trained policy in evaluation mode and selected across multiple random seed runs; and (2) the mean and standard deviation of imagined rewards (i.e., training-time returns) collected during the final 10\% of training, also computed across multiple random seeds.}
  \label{tab:performance-table}
  \begin{center}
    \begin{small}
      \begin{sc}
      \resizebox{\linewidth}{!}{
      \setlength{\tabcolsep}{3.5pt}
        \begin{tabular}{lccccccccc}
          \toprule
\rowcolor{gray!4} 
          Method  & Steps  & Cheetah-Run & Walker-S & Reacher-E & Hopper-Hop & Walker-W & Walker-R & Avg. \\
          \midrule
DreamerV3    & $500\mathrm{k}$ & $689.9$ & $947.8$ & $951.2$ & $245.7$ & $951.5$ & $624.3$ & $735.1$ \\
Dreamer-INFO & $500\mathrm{k}$ & $691.3$ & $934.1$ & $963.4$ & $212.3$ & $904.2$ & $484.7$ & $698.3$ \\
HRSSM        & $500\mathrm{k}$ & $647.9$ & $962.8$ & $868.1$ & $236.7$ & $941.6$ & $515.8$ & $695.5$ \\
DreamerPro   & $500\mathrm{k}$ & $398.0$ & $960.4$ & $964.6$ & $291.6$ & $937.2$ & $527.5$ & $679.9$ \\
R2Dreamer    & $500k$ & $701.1$ & $972.2$ & $970.8$ & $297.9$ & $959.8$ & $673.4$ & $762.5$ \\
\rowcolor{gray!15} 
Ours         & $500k$ & $\mathbf{798.6}$ & $\mathbf{974.7}$ & $\mathbf{985.1}$ & $\mathbf{314.8}$ & $\mathbf{967.2}$ & $\mathbf{694.8}$ & $\mathbf{789.2}$ \\
\midrule
\rowcolor{gray!8} 
\multicolumn{9}{@{}l}{Imagined Rewards} \\
          \midrule
          DreamerV3   & $500k$ & $553.8\pm71.9$ & $939.8\pm1.6$ & $934.6\pm14.9$ & $213.4\pm42.7$ & $925.6\pm27.9$ & $567.2\pm33.1$ & $689.1$ \\
          Dreamer-INFO & $500k$ & $564.0\pm50.8$ & $946.6\pm3.3$ & $955.6\pm16.6$ & $187.7\pm21.2$ & $867.9\pm21.0$ & $343.1\pm38.0$ & $644.1$ \\
          HRSSM        & $500k$ & $525.5\pm96.8$ & $926.5\pm6.8$ & $825.9\pm20.4$ & $203.9\pm38.1$ & $935.3\pm23.6$ & $371.7\pm82.9$ & $631.5$ \\
          DreamerPro   & $500k$ & $428.1\pm62.7$ & $944.2\pm2.5$ & $948.1\pm17.4$ & $269.5\pm35.9$ & $915.8\pm36.2$ & $387.8\pm76.2$ & $648.9$ \\

          R2Dreamer   & $500k$ & $594.8\pm77.3$ & $943.0\pm2.7$ & $917.3\pm18.2$ & $250.4\pm11.7$ & $926.0\pm17.1$ & $583.7\pm14.7$ & $702.5$ \\
          \rowcolor{gray!13}
          Ours        & $500k$ & $\mathbf{687.1\pm34.7}$ & $\mathbf{944.3\pm1.3}$ & $\mathbf{957.7\pm10.5}$ & $\mathbf{282.8\pm14.4}$ & $\mathbf{949.0\pm16.7}$ & $\mathbf{612.7\pm13.9}$ & $\mathbf{738.9}$ \\
          \bottomrule
        \end{tabular}
        }
      \end{sc}
    \end{small}
  \end{center}
  \label{baselineandourstable2}
  \vskip -0.1in
\end{table*}
\subsection{Baselines}

We benchmark our approach against a spectrum of model-based agents, distinguishing between direct structural competitors for dynamic analysis and broader references for asymptotic comparison.
\begin{table}[t]
\centering
\caption{Quantitative analysis of phase-space compactness across tasks. We report the Log Phase Volume ($\sum \log \mathcal{V}$) for four task groups: \textbf{(A)} Cheetah Run, \textbf{(B)} Walker Domain (Walk, Run, and Stand), \textbf{(C)} Reacher Easy, and \textbf{(D)} Hopper Hop. The final row reports the relative reduction $\Delta \mathcal{S}$.}
\label{tab:task_phase_volume_final}
\small
\setlength{\tabcolsep}{4pt}
\begin{tabular}{lcccc}
\toprule
\textbf{Method} & \multicolumn{4}{c}{\textbf{Log Phase Volume} ($\downarrow$)} \\
\cmidrule(lr){2-5}
& \textbf{(A)} & \textbf{(B)} & \textbf{(C)} & \textbf{(D)}\\
\midrule
R2Dreamer & 14.276 & 26.115 & 17.593 & 21.224 \\
\rowcolor{gray!15}
Ours & \textbf{13.227} & \textbf{25.023} & \textbf{16.185} & \textbf{19.439}\\
\midrule
\rowcolor{gray!15}
\textbf{$\Delta \mathcal{S}$} ($\uparrow$) & \textbf{7.35\%} & \textbf{4.18\%} & \textbf{8.00\%}& \textbf{8.41\%}\\
\bottomrule
\end{tabular}
\vspace{-0.1in}
\end{table}

\emph{Primary Comparators.} Our analysis prioritizes R2Dreamer~\citep{morihirar2} as the central baseline. Since our architecture inherits R2Dreamer's robust perceptual backbone but substitutes its generic recurrent transitions with Port-Hamiltonian dynamics, this comparison strictly isolates the gains attributable to physical inductive bias versus implicit and explicit regularization. We additionally compare against DreamerV3, DreamerPro, Dreamer-InfoNCE~\citep{hafner2025mastering,deng2022dreamerpro,oord2018representation}  and HRSSM~\citep{sun2024learning} to benchmark sample efficiency against the state of the art in unstructured, general purpose world models.

\subsection{DMC Results and Phase Space Analysis}
\label{subsec:dmc_results}
\begin{figure}[h] 
  \centering
  \begin{minipage}{0.48\textwidth}
    \centering
    \includegraphics[width=\linewidth]{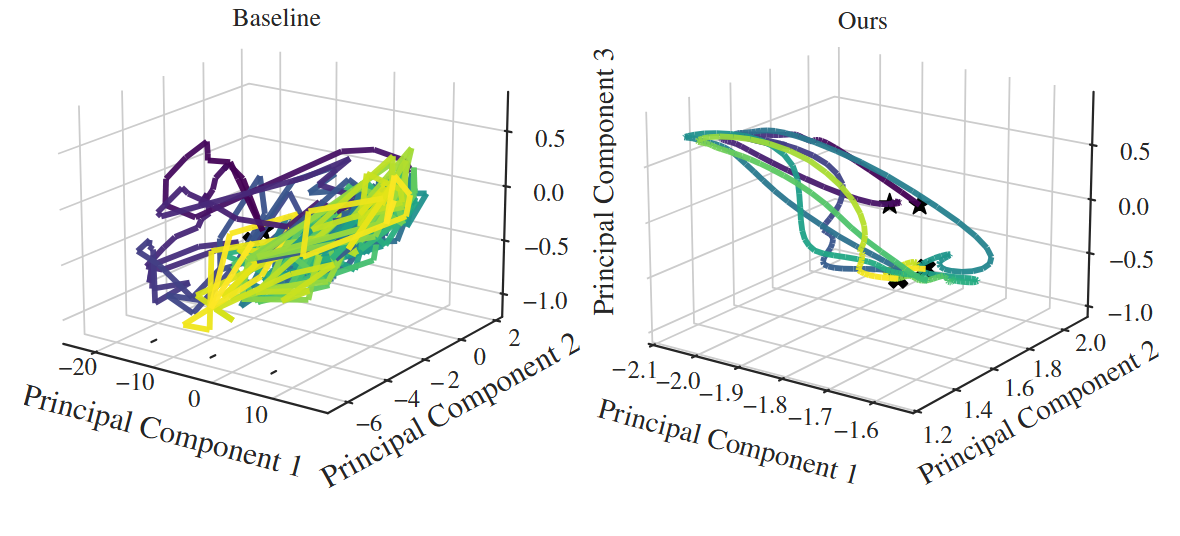}
  \end{minipage}
  \begin{minipage}{0.48\textwidth}
    \centering
    \includegraphics[width=\linewidth]{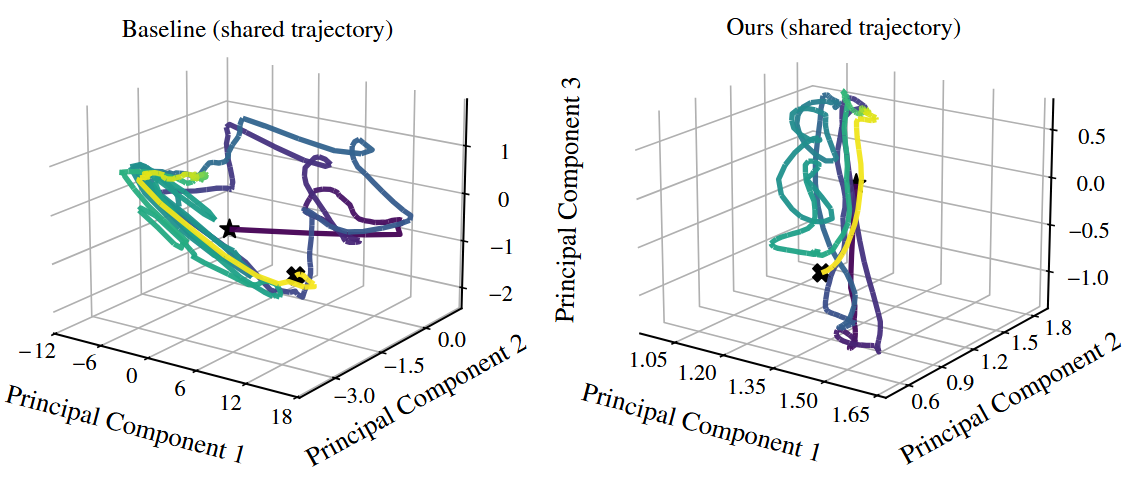}
  \end{minipage}

  \vspace{5pt} 
  \caption{\textbf{Latent phase space analysis.} PH-RSSM yields a more compact projected PH phase space than the baseline. On a shared input trajectory, its projected latents remain more bounded, supporting latent stability gains without implying a closed loop boundedness guarantee.}
  \label{P3D}
\end{figure}

To address Question 1, we employ two critical metrics. We first evaluate the alignment between task returns accumulated during latent imagination and those from real world evaluation. While baseline models achieve strong convergence, PH-RSSM biases imagined trajectories toward PH-regularized latent dynamics, reducing the gap between imagined rewards and actual performance in our experiments (Table~\ref{tab:performance-table}). Second, we quantify latent compactness using the Log Phase Volume metric derived through PCA (Table~\ref{tab:task_phase_volume_final}). We observe that our physical regularizer is associated with a more compact projected PH phase space (see  Figure~\ref{P3D}). The ability to maintain superior task performance within a significantly reduced projected volume confirms that our architecture distills a robust internal physical representation rather than relying on high entropy statistical correlations.

\subsection{Analysis of Physical Fidelity and Behavioral Efficiency}
\label{subsec:analysis}
We investigate whether PH-RSSM internalizes the thermodynamic structure of the environment through two complementary evaluation regimes. 

\begin{figure*}[h]
  \centering
  \includegraphics[width=0.95\textwidth]{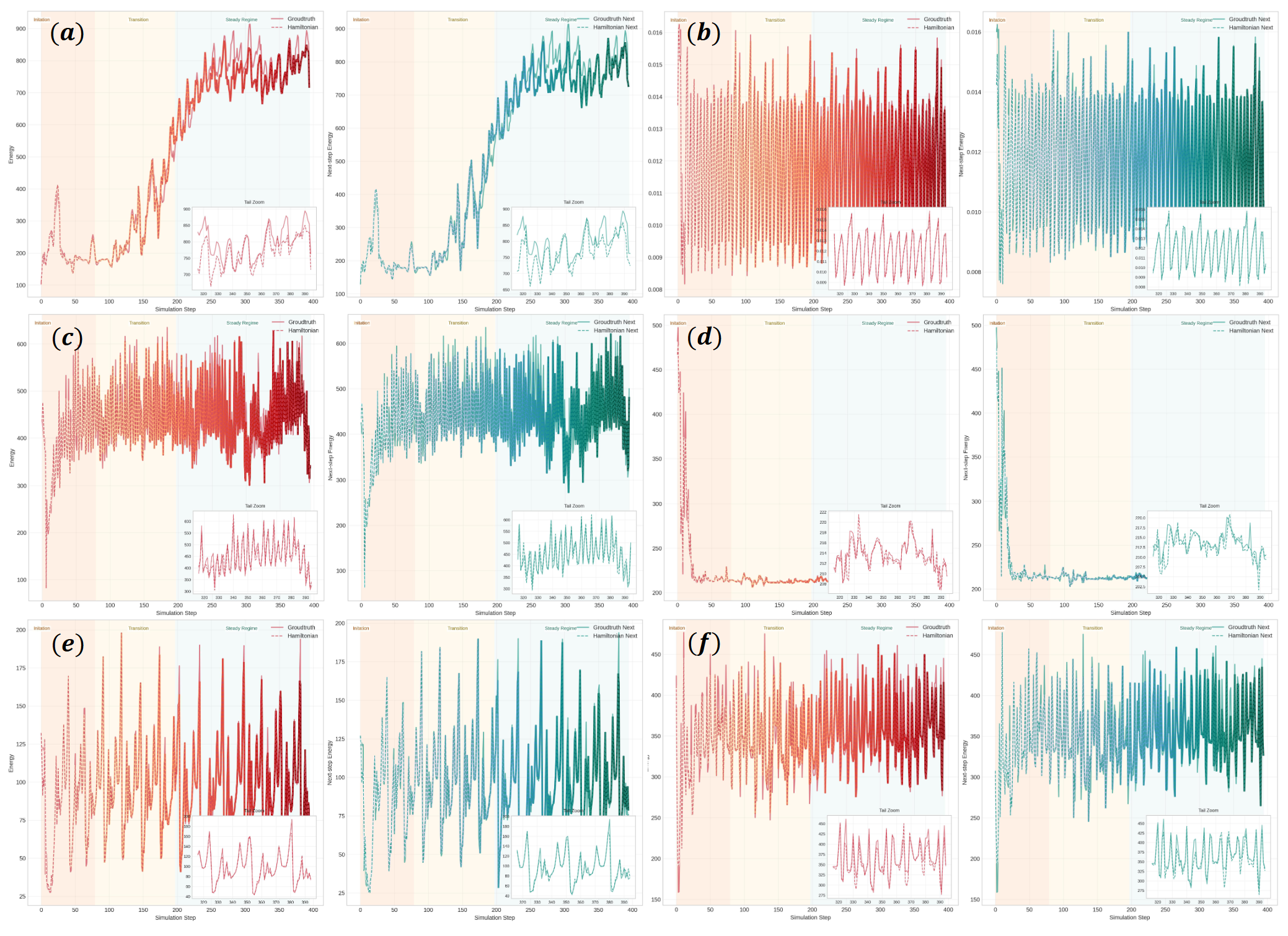}
  \caption{
    \textbf{Explicit energy alignment.} Predicted Hamiltonians are compared with MuJoCo ground truth mechanical energy across six DMC tasks. Each panel reports current-state and post-action energy predictions, with an inset highlighting the tail segment.
}
  \label{PhysicsVis}
  \vskip -0.1in
\end{figure*}

\textbf{Fidelity of Learned Energy Dynamics.} To evaluate whether the learned energy representations capture physically meaningful structure, we analyze their alignment with ground truth mechanical energy from the MuJoCo engine. As shown in Figure~\ref{PhysicsVis}, the explicit energy world model tracks ground truth mechanical energy with high temporal coherence when grounded in proprioceptive kinematic observations. This correlation suggests that the energy model captures meaningful energy fluctuations rather than relying solely on visual pattern memorization, supporting its use for energy aware policy regularization.

\textbf{Evaluation Metrics.} To quantify the policy's performance in terms of energy efficiency and motion smoothness, we define two primary metrics based on the MuJoCo physics engine's internal states:

\begin{enumerate}
    \item \textbf{Total Energy Consumption ($\mathcal{E}$):} We evaluate energy consumption by integrating absolute mechanical work and a weighted control effort proxy (since Joule heating in electric motors scales with squared current, and thus $P_{heat} \propto \tau^2$). Let $\boldsymbol{\tau} \in \mathbb{R}^{n_{act}}$ be the actuator force (\texttt{actuator\_force}) and $\dot{\mathbf{q}} \in \mathbb{R}^{n_{dof}}$ be the joint velocities (\texttt{qvel}). The total energy proxy over an episode of $T$ steps is \citep{seok2014design,yang2022fast}:
    \begin{equation}
        \mathcal{E}_{\mathrm{proxy}} = \alpha\sum_{t=1}^{T}\sum_{i=1}^{n_{act}} |\tau_{i,t}\dot{q}_{\mathcal{M}(i),t}|\Delta t
        + \beta\sum_{t=1}^{T}\sum_{i=1}^{n_{act}}\tau_{i,t}^{2}\Delta t
    \end{equation}
    where $\mathcal{M}(i)$ maps actuators to DoFs. Fixed $\alpha,\beta$ are shared by all methods; cross-task averages use normalized values.

    \item \textbf{Mean Squared Jerk ($\mathcal{J}$):} Smoothness is measured by the temporal derivative of joint accelerations $\ddot{\mathbf{q}}$ (\texttt{qacc}). The average jerk across all $n_{dof}$ joints is defined as \citep{balasubramanian2015analysis}:
    \begin{equation}
        \mathcal{J} = \frac{1}{(T-1)n_{dof}} \sum_{t=2}^{T} \sum_{j=1}^{n_{dof}} \left( \frac{\ddot{q}_{j,t} - \ddot{q}_{j,t-1}}{\Delta t} \right)^2
    \end{equation}
\end{enumerate}

Crucially, by internalizing the Port-Hamiltonian inductive bias, our energy-guided Actor-Critic optimization explicitly minimizes these two complementary metrics. As demonstrated in our quantitative analysis (Table \ref{tab:energy_smoothness_eval}), compared to the unstructured baseline, PH-RSSM achieves relative improvements of up to 7.80\% in Total Energy Consumption ($\mathcal{E}$) and 9.38\% in Mean Squared Jerk ($\mathcal{J}$), computed per task before averaging. These results suggest that penalizing inferred action driven energy variation and smoothness violations can steer the agent toward lower energy and smoother control strategies without degrading task rewards.

\begin{table}[H]
\centering
\small
\setlength{\tabcolsep}{8pt}
\begin{tabular}{lcc}
\toprule
\textbf{Method} & \textbf{TEC} ($\mathcal{E} \downarrow$) & \textbf{MSJ} ($\mathcal{J} \downarrow$) \\
\midrule
HRSSM      & 125.43 & 45.12 \\
DreamerV3  & 132.89 & 48.64 \\
DreamerPro & 128.36 & 46.87 \\
R2Dreamer  & 122.10 & 44.05 \\
\midrule
\rowcolor{gray!15}
\textit{w/o} Implicit & 117.52 & 40.19 \\
\rowcolor{gray!15}
\textit{w/o} Explicit & 121.84 & 43.63 \\
\rowcolor{gray!15}
Ours & \textbf{112.58} & \textbf{39.92} \\
\bottomrule
\end{tabular}
\caption{Quantitative analysis of control efficiency and trajectory smoothness. We evaluate the performance using two proposed metrics: Total Energy Consumption (TEC) and Mean Squared Jerk (MSJ). TEC is a fixed weighted proxy of mechanical work and thermal control effort, while MSJ quantifies smoothness via acceleration differentials. Aggregation first computes each task's relative change or normalized score, then averages across tasks.}
\label{tab:energy_smoothness_eval}
\end{table}
\section{Discussion}
\textbf{Limitations}
PH-RSSM improves physical regularity and control efficiency, but its return gains are modest and its PH constraints remain auxiliary inductive biases rather than closed-loop guarantees. The explicit energy branch also relies on proprioceptive and simulator-derived signals, so it does not yet demonstrate purely pixel-based physical discovery.

\textbf{Future Work}
Future work will ground latent Hamiltonians in measurable quantities, refine momentum estimation with temporal context, and explore more accurate symplectic integrators. We also plan broader validation across more complex settings and, eventually, real robotic systems.

\section{Conclusion}
We introduced PH-RSSM, a Port-Hamiltonian module for RSSM latents that encourages more compact and physically structured representations. Across our benchmarks, it improves latent compactness, energy efficiency, and average return over R2Dreamer.


\clearpage
\bibliographystyle{plainnat}
\bibliography{reference}

\begin{thebibliography}{38}
\providecommand{\natexlab}[1]{#1}
\providecommand{\url}[1]{\texttt{#1}}
\expandafter\ifx\csname urlstyle\endcsname\relax
  \providecommand{\doi}[1]{doi: #1}\else
  \providecommand{\doi}{doi: \begingroup \urlstyle{rm}\Url}\fi

\bibitem[Ai et~al.(2025)Ai, Tian, Shi, Wang, Pfaff, Tan, Christensen, Su, Wu, and Li]{ai2025review}
Bo~Ai, Stephen Tian, Haochen Shi, Yixuan Wang, Tobias Pfaff, Cheston Tan, Henrik~I. Christensen, Hao Su, Jiajun Wu, and Yunzhu Li.
\newblock A review of learning-based dynamics models for robotic manipulation.
\newblock \emph{Science Robotics}, 10\penalty0 (106):\penalty0 eadt1497, 2025.

\bibitem[Balasubramanian et~al.(2015)Balasubramanian, Melendez-Calderon, Roby-Brami, and Burdet]{balasubramanian2015analysis}
Sivakumar Balasubramanian, Alejandro Melendez-Calderon, Agnes Roby-Brami, and Etienne Burdet.
\newblock On the analysis of movement smoothness.
\newblock \emph{Journal of NeuroEngineering and Rehabilitation}, 12\penalty0 (1):\penalty0 112, 2015.

\bibitem[Bastek et~al.(2025)Bastek, Sun, and Kochmann]{bastek2025physics}
Jan{-}Hendrik Bastek, WaiChing Sun, and Dennis~M. Kochmann.
\newblock Physics-informed diffusion models.
\newblock In \emph{International Conference on Learning Representations (ICLR)}, 2025.

\bibitem[Brooks et~al.(2024)Brooks, Peebles, Holmes, DePue, Guo, Jing, Schnurr, Taylor, Luhman, Luhman, Ng, Wang, and Ramesh]{brooks2024sora}
Tim Brooks, Bill Peebles, Connor Holmes, Will DePue, Yufei Guo, Li~Jing, David Schnurr, Joe Taylor, Troy Luhman, Eric Luhman, Clarence Ng, Ricky Wang, and Aditya Ramesh.
\newblock Video generation models as world simulators, 2024.
\newblock URL \url{https://openai.com/index/video-generation-models-as-world-simulators/}.
\newblock OpenAI technical report.

\bibitem[de~Avila~Belbute{-}Peres et~al.(2018)de~Avila~Belbute{-}Peres, Smith, Allen, Tenenbaum, and Kolter]{de2018end}
Filipe de~Avila~Belbute{-}Peres, Kevin~A. Smith, Kelsey~R. Allen, Josh Tenenbaum, and J.~Zico Kolter.
\newblock End-to-end differentiable physics for learning and control.
\newblock In \emph{Advances in Neural Information Processing Systems}, volume~31, pages 7178--7189, 2018.

\bibitem[Deng et~al.(2022)Deng, Jang, and Ahn]{deng2022dreamerpro}
Fei Deng, Ingook Jang, and Sungjin Ahn.
\newblock {DreamerPro}: Reconstruction-free model-based reinforcement learning with prototypical representations.
\newblock In \emph{Proceedings of the 39th International Conference on Machine Learning}, volume 162 of \emph{Proceedings of Machine Learning Research}, pages 4956--4975. PMLR, 2022.

\bibitem[Dongare et~al.(2025)Dongare, Kharel, Samuel, and Zhou]{dongare2025transdreamerv3}
Shruti~Sadanand Dongare, Amun Kharel, Jonathan Samuel, and Xiaona Zhou.
\newblock {T}rans{D}reamer{V}3: Implanting {T}ransformer in {D}reamer{V}3.
\newblock \emph{arXiv preprint arXiv:2506.17103}, 2025.

\bibitem[Ferraro et~al.(2025)Ferraro, Mazzaglia, Verbelen, and Dhoedt]{ferraro2025focus}
Stefano Ferraro, Pietro Mazzaglia, Tim Verbelen, and Bart Dhoedt.
\newblock {FOCUS}: object-centric world models for robotic manipulation.
\newblock \emph{Frontiers in Neurorobotics}, 19:\penalty0 1585386, 2025.

\bibitem[{Genesis Authors}(2024)]{zhou2024genesis}
{Genesis Authors}.
\newblock {G}enesis: A generative and universal physics engine for robotics and beyond, December 2024.
\newblock URL \url{https://github.com/Genesis-Embodied-AI/Genesis}.

\bibitem[Greydanus et~al.(2019)Greydanus, Dzamba, and Yosinski]{greydanus2019hamiltonian}
Samuel Greydanus, Misko Dzamba, and Jason Yosinski.
\newblock {H}amiltonian neural networks.
\newblock In \emph{Advances in Neural Information Processing Systems}, volume~32, pages 15353--15363, 2019.

\bibitem[Guo et~al.(2025)Guo, Huo, Shi, Song, Zhang, and Zhao]{guo2025t2vphysbench}
Xuyang Guo, Jiayan Huo, Zhenmei Shi, Zhao Song, Jiahao Zhang, and Jiale Zhao.
\newblock {T}2{V}{P}hys{B}ench: {A} first-principles benchmark for physical consistency in text-to-video generation.
\newblock \emph{arXiv preprint arXiv:2505.00337}, 2025.

\bibitem[Hafner et~al.(2019)Hafner, Lillicrap, Fischer, Villegas, Ha, Lee, and Davidson]{hafner2019learning}
Danijar Hafner, Timothy~P. Lillicrap, Ian Fischer, Ruben Villegas, David Ha, Honglak Lee, and James Davidson.
\newblock Learning latent dynamics for planning from pixels.
\newblock In \emph{Proceedings of the 36th International Conference on Machine Learning}, volume~97 of \emph{Proceedings of Machine Learning Research}, pages 2555--2565. PMLR, 2019.

\bibitem[Hafner et~al.(2020)Hafner, Lillicrap, Ba, and Norouzi]{hafner2020dream}
Danijar Hafner, Timothy~P. Lillicrap, Jimmy Ba, and Mohammad Norouzi.
\newblock Dream to control: Learning behaviors by latent imagination.
\newblock In \emph{International Conference on Learning Representations (ICLR)}, 2020.

\bibitem[Hafner et~al.(2021)Hafner, Lillicrap, Norouzi, and Ba]{hafner2021mastering}
Danijar Hafner, Timothy~P. Lillicrap, Mohammad Norouzi, and Jimmy Ba.
\newblock Mastering atari with discrete world models.
\newblock In \emph{International Conference on Learning Representations (ICLR)}, 2021.

\bibitem[Hafner et~al.(2025{\natexlab{a}})Hafner, Pasukonis, Ba, and Lillicrap]{hafner2025mastering}
Danijar Hafner, Jurgis Pasukonis, Jimmy Ba, and Timothy~P. Lillicrap.
\newblock Mastering diverse control tasks through world models.
\newblock \emph{Nature}, 640\penalty0 (8059):\penalty0 647--653, 2025{\natexlab{a}}.

\bibitem[Hafner et~al.(2025{\natexlab{b}})Hafner, Yan, and Lillicrap]{hafner2025training}
Danijar Hafner, Wilson Yan, and Timothy~P. Lillicrap.
\newblock Training agents inside of scalable world models.
\newblock \emph{arXiv preprint arXiv:2509.24527}, 2025{\natexlab{b}}.

\bibitem[Huang et~al.(2024)Huang, Ji, Xia, Zhang, and Yang]{huang2024safedreamer}
Weidong Huang, Jiaming Ji, Chunhe Xia, Borong Zhang, and Yaodong Yang.
\newblock {S}afe{D}reamer: Safe reinforcement learning with world models.
\newblock In \emph{International Conference on Learning Representations (ICLR)}, 2024.

\bibitem[Kang et~al.(2025)Kang, Yue, Lu, Lin, Zhao, Wang, Huang, and Feng]{kang2025far}
Bingyi Kang, Yang Yue, Rui Lu, Zhijie Lin, Yang Zhao, Kaixin Wang, Gao Huang, and Jiashi Feng.
\newblock How far is video generation from world model: {A} physical law perspective.
\newblock In \emph{Proceedings of the 42nd International Conference on Machine Learning}, volume 267 of \emph{Proceedings of Machine Learning Research}, pages 28991--29017. PMLR, 2025.

\bibitem[Levy et~al.(2025)Levy, Westenbroek, and Fridovich{-}Keil]{levy2025learning}
Jacob Levy, Tyler Westenbroek, and David Fridovich{-}Keil.
\newblock Learning to walk from three minutes of real-world data with semi-structured dynamics models.
\newblock In \emph{Proceedings of The 8th Conference on Robot Learning}, volume 270 of \emph{Proceedings of Machine Learning Research}, pages 2061--2079. PMLR, 2025.

\bibitem[Li et~al.(2025)Li, Zhao, Yu, Du, Zou, Hu, and Xu]{li2025pin}
Wenxuan Li, Hang Zhao, Zhiyuan Yu, Yu~Du, Qin Zou, Ruizhen Hu, and Kai Xu.
\newblock {PIN-WM}: Learning physics-informed world models for non-prehensile manipulation.
\newblock In \emph{Robotics: Science and Systems XXI}, 2025.

\bibitem[Liu et~al.(2024)Liu, Ren, Gupta, and Wang]{liu2024physgen}
Shaowei Liu, Zhongzheng Ren, Saurabh Gupta, and Shenlong Wang.
\newblock {P}hys{G}en: Rigid-body physics-grounded image-to-video generation.
\newblock In \emph{European Conference on Computer Vision (ECCV)}, volume 15140, pages 360--378, 2024.

\bibitem[Lutter et~al.(2019)Lutter, Ritter, and Peters]{Lutter2019RP19}
Michael Lutter, Christian Ritter, and Jan Peters.
\newblock Deep {Lagrangian} networks: Using physics as model prior for deep learning.
\newblock In \emph{International Conference on Learning Representations (ICLR)}. OpenReview.net, 2019.

\bibitem[Morihira et~al.(2026)Morihira, Nahar, Bharadwaj, Kato, Hayashi, and Harada]{morihirar2}
Naoki Morihira, Amal Nahar, Kartik Bharadwaj, Yasuhiro Kato, Akinobu Hayashi, and Tatsuya Harada.
\newblock {R2-Dreamer}: Redundancy-reduced world models without decoders or augmentation.
\newblock In \emph{International Conference on Learning Representations (ICLR)}, 2026.

\bibitem[Mosbach et~al.(2025)Mosbach, Ewertz, Villar{-}Corrales, and Behnke]{mosbach2025sold}
Malte Mosbach, Jan~Niklas Ewertz, Angel Villar{-}Corrales, and Sven Behnke.
\newblock {SOLD}: Slot object-centric latent dynamics models for relational manipulation learning from pixels.
\newblock In \emph{Proceedings of the 42nd International Conference on Machine Learning}, volume 267 of \emph{Proceedings of Machine Learning Research}, pages 44911--44935. PMLR, 2025.

\bibitem[Motamed et~al.(2025)Motamed, Culp, Swersky, Jaini, and Geirhos]{motamed2025generative}
Saman Motamed, Laura Culp, Kevin Swersky, Priyank Jaini, and Robert Geirhos.
\newblock Do generative video models understand physical principles?
\newblock \emph{arXiv preprint arXiv:2501.09038}, 2025.

\bibitem[Oord et~al.(2018)Oord, Li, and Vinyals]{oord2018representation}
Aaron van~den Oord, Yazhe Li, and Oriol Vinyals.
\newblock Representation learning with contrastive predictive coding.
\newblock \emph{arXiv preprint arXiv:1807.03748}, 2018.

\bibitem[Ramesh and Ravindran(2023)]{ramesh2023physics}
Adithya Ramesh and Balaraman Ravindran.
\newblock Physics-informed model-based reinforcement learning.
\newblock In \emph{Proceedings of The 5th Annual Learning for Dynamics and Control Conference}, volume 211 of \emph{Proceedings of Machine Learning Research}, pages 26--37. PMLR, 2023.

\bibitem[Seok et~al.(2015)Seok, Wang, Chuah, Hyun, Lee, Otten, Lang, and Kim]{seok2014design}
Sangok Seok, Albert Wang, Meng~Yee Chuah, Dong~Jin Hyun, Jongwoo Lee, David~M Otten, Jeffrey~H Lang, and Sangbae Kim.
\newblock Design principles for energy-efficient legged locomotion and implementation on the {MIT} cheetah robot.
\newblock \emph{IEEE/ASME Transactions on Mechatronics}, 20\penalty0 (3):\penalty0 1117--1129, 2015.

\bibitem[Shang et~al.(2025)Shang, Zhang, Tang, Jin, Gao, Wu, and Li]{shang2025roboscape}
Yu~Shang, Xin Zhang, Yinzhou Tang, Lei Jin, Chen Gao, Wei Wu, and Yong Li.
\newblock Roboscape: Physics-informed embodied world model.
\newblock \emph{arXiv preprint arXiv:2506.23135}, 2025.

\bibitem[Shi et~al.(2023)Shi, Ding, Cao, Yao, Liu, and Li]{shi2023learning}
Hongzhi Shi, Jingtao Ding, Yufan Cao, Quanming Yao, Li~Liu, and Yong Li.
\newblock Learning symbolic models for graph-structured physical mechanism.
\newblock In \emph{International Conference on Learning Representations (ICLR)}, 2023.

\bibitem[Sun et~al.(2024)Sun, Zang, Li, and Islam]{sun2024learning}
Ruixiang Sun, Hongyu Zang, Xin Li, and Riashat Islam.
\newblock Learning latent dynamic robust representations for world models.
\newblock In \emph{Proceedings of the 41st International Conference on Machine Learning}, volume 235 of \emph{Proceedings of Machine Learning Research}, pages 47234--47260. PMLR, 2024.

\bibitem[Tassa et~al.(2018)Tassa, Doron, Muldal, Erez, Li, de~Las~Casas, Budden, Abdolmaleki, Merel, Lefrancq, Lillicrap, and Riedmiller]{tassa2018deepmind}
Yuval Tassa, Yotam Doron, Alistair Muldal, Tom Erez, Yazhe Li, Diego de~Las~Casas, David Budden, Abbas Abdolmaleki, Josh Merel, Andrew Lefrancq, Timothy~P. Lillicrap, and Martin~A. Riedmiller.
\newblock {D}eep{M}ind control suite.
\newblock \emph{arXiv preprint arXiv:1801.00690}, 2018.

\bibitem[Toth et~al.(2020)Toth, Rezende, Jaegle, Racani{\`{e}}re, Botev, and Higgins]{toth2020hamiltonian}
Peter Toth, Danilo~J. Rezende, Andrew Jaegle, S{\'{e}}bastien Racani{\`{e}}re, Aleksandar Botev, and Irina Higgins.
\newblock {H}amiltonian generative networks.
\newblock In \emph{International Conference on Learning Representations (ICLR)}, 2020.

\bibitem[Yang et~al.(2020)Yang, He, and Zhu]{yang2020learning}
Shuqi Yang, Xingzhe He, and Bo~Zhu.
\newblock Learning physical constraints with neural projections.
\newblock In \emph{Advances in Neural Information Processing Systems}, volume~33, 2020.

\bibitem[Yang et~al.(2022)Yang, Zhang, Coumans, Tan, and Boots]{yang2022fast}
Yuxiang Yang, Tingnan Zhang, Erwin Coumans, Jie Tan, and Byron Boots.
\newblock Fast and efficient locomotion via learned gait transitions.
\newblock In \emph{Proceedings of the 5th Conference on Robot Learning}, volume 164 of \emph{Proceedings of Machine Learning Research}, pages 773--783. PMLR, 2022.

\bibitem[Ying et~al.(2023)Ying, Zhou, Hao, Su, Liu, Yan, and Zhu]{ying2023task}
Chengyang Ying, Xinning Zhou, Zhongkai Hao, Hang Su, Songming Liu, Dong Yan, and Jun Zhu.
\newblock Task aware dreamer for task generalization in reinforcement learning.
\newblock \emph{arXiv preprint arXiv:2303.05092}, 2023.

\bibitem[Zhang et~al.(2023)Zhang, Wang, Sun, Yuan, and Huang]{zhang2023storm}
Weipu Zhang, Gang Wang, Jian Sun, Yetian Yuan, and Gao Huang.
\newblock {STORM}: Efficient stochastic transformer based world models for reinforcement learning.
\newblock In \emph{Advances in Neural Information Processing Systems}, volume~36, pages 27147--27166, 2023.

\bibitem[Zheng et~al.(2025)Zheng, Huang, Liu, Zou, He, Zhang, Zhang, He, Zheng, Qiao, and Liu]{zheng2025vbench}
Dian Zheng, Ziqi Huang, Hongbo Liu, Kai Zou, Yinan He, Fan Zhang, Yuanhan Zhang, Jingwen He, Wei{-}Shi Zheng, Yu~Qiao, and Ziwei Liu.
\newblock {V}{B}ench-2.0: Advancing video generation benchmark suite for intrinsic faithfulness.
\newblock \emph{arXiv preprint arXiv:2503.21755}, 2025.

\end{thebibliography}
\end{document}